\documentclass{Interspeech}

\interspeechcameraready

\title{NIRANTAR: Continual Learning with New Languages and Domains on Real-world Speech Data}

\author[affiliation={1}]{Tahir}{Javed}
\author[affiliation={1}]{Kaushal}{Bhogale}
\author[affiliation={1}]{Mitesh M.}{Khapra}

\affiliation{}{AI4Bharat, Indian Institute of Technology Madras}{India}
\email{\{tahir, cs22d006, miteshk\}@cse.iitm.ac.in}


\usepackage{comment}
\usepackage{hyperref}
\usepackage{url}

\usepackage{graphicx}
\usepackage{booktabs}
\usepackage[table]{xcolor}
\usepackage{multirow}
\usepackage{colortbl}
\usepackage{soul}
\usepackage{pifont}
\usepackage{color}
\usepackage{subcaption}
\usepackage{longtable}
\usepackage{tablefootnote}

\newcommand{\framework}{\textsc{Nirantar}}
\newcommand{\dataset}{Nirantar}

\newcommand{\cmark}{\textcolor{green}{\textbf{\ding{51}}}}%
\newcommand{\xmark}{\textcolor{red}{\textbf{\ding{55}}}}%

\begin{document}

\maketitle

\begin{abstract}
    
    We introduce Nirantar\footnote{\label{fn:nirantar}Nirantar means \textit{continual} in Hindi. \par \url{https://github.com/AI4Bharat/Nirantar}}, a comprehensive framework for evaluating continual learning (CL) in multilingual and multi-domain ASR. Designed to reflect real-world CL challenges, Nirantar leverages data collected incrementally across 22 languages and 208 districts in India through natural episodes. This enables evaluation across Language-Incremental (LIL), Domain-Incremental (DIL), and the novel Language-Incremental Domain-Incremental Learning (LIDIL) scenarios. Unlike prior work that relies on simulated episodes, Nirantar presents dynamic, non-uniform language and domain shifts, making it an ideal testbed for CL research. With 3250 hours of human-transcribed speech, including 1720 hours newly introduced in this work, our framework enables systematic benchmarking of CL methods. We evaluate existing approaches and demonstrate that no single method performs consistently well, underscoring the need for more robust CL strategies. 
\end{abstract}

\section{Introduction}
\label{sec:intro}

There is a growing trend towards training massive multilingual speech models on large datasets \cite{commonvoice, voxpopuli}
aggregated across multiple languages \cite{m-ctc-t, whisper, google-usm}. Given the high computational demands, continual training is essential as new datasets covering additional languages, domains, or demographics are introduced over time \cite{commonvoice, springx}. To address this, continual learning (CL) techniques have emerged \cite{cl_survey, cl_survey_2}, allowing efficient model updates while preserving prior knowledge across \textit{instance incremental learning}, \textit{task incremental learning}, and \textit{domain incremental learning}. However, most CL datasets \cite{permuted-mnist, split-mnist}
, are synthetically created, lacking natural episodes, making them unsuitable for real-world CL evaluation. More recent real-world benchmarks \cite{clear, visual_decathlon, clif}
focus on either task or domain incremental learning but fail to address both simultaneously.

\begin{figure}[!t]
    \centering
    \includegraphics[width=\linewidth]{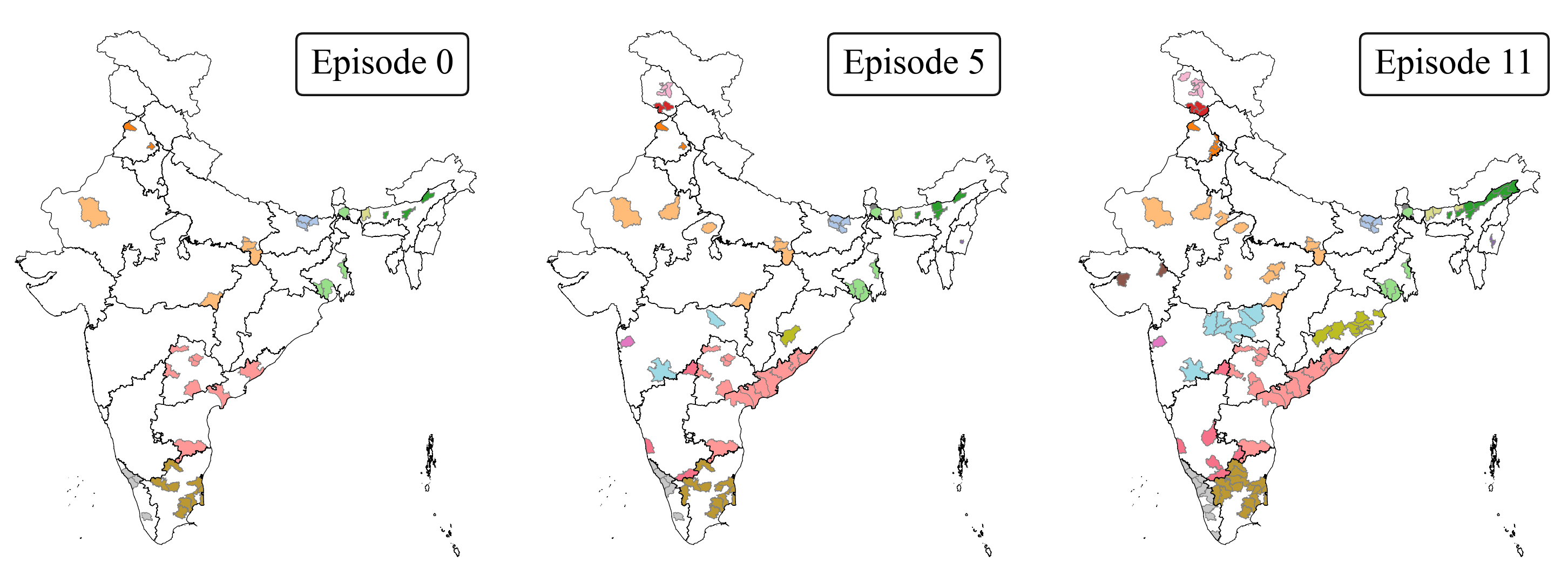}
    \caption{Illustration of Language-Incremental Domain-Incremental Learning: A practical scenario showing the addition of both new languages and domains in each episode of speech data collection.}
    \label{fig:motivation}
\end{figure}

In this work, we release a real-world CL playground by building on the IndicVoices \cite{indicvoices} initiative. 
We extend this effort by expanding coverage, increasing data volume, and introducing new domains for a more comprehensive multilingual dataset covering 22 low-resource Indian languages and 400 districts. Our data collection happens in batches, with each batch targeting specific districts and languages. Each district is treated as a distinct domain due to its unique vocabulary, accents, and local interests. For instance, speakers from Srinagar may discuss snow-capped mountains, while those from Assam may talk about tea plantations. From each district, 20 to 50 hours of read, extempore, and conversational speech is collected, covering diverse topics such as farming, education, tourism, politics, etc.

The episodic nature of this data collection, with periodic gaps between batches, creates a natural setting for continual learning (CL). Leveraging this, we introduce Nirantar, a CL framework designed for three scenarios: Language-Incremental (LIL), Domain-Incremental (DIL), and the novel Language-Incremental Domain-Incremental Learning (LIDIL) introduced as a part of this work (See Figure \ref{fig:motivation}). Nirantar consists of 3,250 hours of human-transcribed speech, including 1,530 hours from IndicVoices and 1,720 newly collected hours as a part of this work. The training data is divided into 12 episodes, each introducing new languages, domains, or both. The evaluation set includes 15 minutes of diverse speech per domain-language pair, continuously updated as new data is collected, making it a live, evolving benchmark for CL research. Nirantar covers 22 languages from 4 language families, spanning medium-resource (e.g., Tamil, Bengali), low-resource (e.g., Marathi, Urdu), and extremely low-resource (e.g., Sindhi, Bodo) languages. The insights from Nirantar would thus be relevant to other low-resource language groups and diverse language families.

We evaluate several CL approaches on Nirantar, including replay-based methods like Experience Replay \cite{experience-replay} and regularization-based methods such as Elastic Weight Consolidation \cite{DBLP:conf/aaai/LiuYW21} and Memory-aware Synapse \cite{memory-aware-synapse}. These methods exhibit varying performance across the three CL scenarios, underscoring the need for more robust techniques that perform consistently in multilingual and multidomain settings. Additionally, we find that architecture-based CL methods, which require adding parameters to the backbone model, are impractical in real-world scenarios. For instance, supporting 22 languages and 208 domains in Nirantar would necessitate adding a new adapter per language and domain, leading to excessive model complexity and scalability issues. This observation raises concerns about the feasibility of such methods for large-scale CL applications. To facilitate further research, we have made all code, data, and models available\footref{fn:nirantar} under the CC-BY-4.0 license.

\section{Related work}

Continual Learning (CL) in ASR has been explored mainly in Language-Incremental and Domain-Incremental Learning \cite{DBLP:journals/natmi/VenTT22}. Prior work includes domain-specific ASR sub-models \cite{DBLP:conf/interspeech/SadhuH20} and monolingual hybrid CTC-transformer adaptation \cite{DBLP:conf/interspeech/ChangLL21}, both focusing on domain-incremental setups. CL-MASR \cite{DBLP:journals/corr/abs-2310-16931} examines CL strategies in a multilingual setting, emphasizing language-incremental learning. However, real-world scenarios remain similar to ours remain underexplored.  While the NIC setting \cite{core50, DBLP:journals/corr/abs-2404-06859} addresses new instances and classes, our work  is the first to provide a robust framework for multilingual and multi-domain continual learning for ASR. Figure 3 compares \dataset{} to other ASR datasets and shows that none of the existing datasets support all the 3 scenarios considered in this work.

Existing CL approaches fall into three categories \cite{DBLP:journals/corr/abs-2302-00487}. 
First, regularization-based methods, such as Elastic Weight Consolidation (EWC) \cite{DBLP:conf/aaai/LiuYW21} and Memory-aware Synapses (MAS) \cite{memory-aware-synapse}, limit large weight updates to retain prior knowledge. Second, replay-based approaches like Experience Replay (ER) \cite{experience-replay} and its variants, including Dark Experience Replay (DER) \cite{DER} and A-GEM \cite{AGEM}, store past examples to mitigate forgetting. Third, architecture-based methods, such as Adapters \cite{DBLP:conf/icassp/Eeckth23a}, Progressive Neural Networks (PNNs) \cite{rusu2016progressive} and PackNet \cite{packtnet}, allocate dedicated parameters for new tasks. 
We evaluate a representative set of these approaches on Nirantar and find that no single method performs consistently well.

\section{\framework: CL on Real-World Data}

This section introduces Nirantar, a playground for continual learning in ASR with new languages and domains. We now introduce definitions which will be used through the paper.

\begingroup
\setlength{\tabcolsep}{4pt} 

\begin{table}[!t]
\scriptsize
\centering
\caption{Table comparing different publicly available datasets and their usability in different CL scenarios. (Tr = Transcription, FA = Force Aligned, PL = Pseudo Labelled, M = Manual, \#L = Languages, \#D = Domains)}
\begin{tabular}{@{}l|l@{}rrll|c@{}c@{}c@{}}
\toprule
\multirow{2}{*}{\textbf{Dataset}} & \multirow{2}{*}{\textbf{\#L}} & \multirow{2}{*}{\textbf{\#D}} & \multirow{2}{*}{\textbf{\#H}} & \multirow{2}{*}{\begin{tabular}[c]{@{}c@{}}\textbf{Audio}\\\textbf{Source}\end{tabular}} & \multirow{2}{*}{\textbf{Tr}} & \multicolumn{3}{c}{\textbf{Scenario}} \\
 &  &  &  & \multicolumn{1}{c}{} & \multicolumn{1}{c|}{} & \multicolumn{1}{c}{LIL} & \multicolumn{1}{@{}c@{}}{DIL} & \multicolumn{1}{c@{}}{LIDIL} \\
 \midrule
LibriSpeech  & 1 & - & 1000 & Audiobooks & FA & \xmark & \xmark & \xmark \\
GigaSpeech  & 1 & 23 & 10000 & YouTube & FA & \xmark & \cmark & \xmark \\
VoxPopuli & 16 & - & 1800 & \begin{tabular}[c]{@{}l@{}}Parliament\\Recordings\end{tabular} & FA & \cmark & \xmark & \xmark \\
TED-LIUM & 1 & - & 452 & TED talks & FA & \xmark & \xmark & \xmark \\
Spoken Wikipedia & 3 & - & 1005 & \begin{tabular}[c]{@{}l@{}}Crowd\\sourcing\end{tabular} & FA & \cmark & \xmark & \xmark \\
Multilingual-TEDx  & 8 & - & 765 & TED Talks & FA & \cmark & \xmark & \xmark \\
\begin{tabular}[c]{@{}l@{}}Multilingual\\LibriSpeech \end{tabular}  & 8 & - & 44500 & Audiobooks & FA & \cmark & \xmark & \xmark \\
GigaSpeech 2 & 3 & - & 22015 & YouTube & PL & \cmark & \xmark & \xmark \\
Switchboard  & 1 & - & 260 & Human & M & \xmark & \xmark & \xmark \\
CommonVoice & 131 & - & 21594 & Human & M & \cmark & \xmark & \xmark \\

FLEURS  & 102 & - & 1400 & Human & M & \cmark & \xmark & \xmark \\
MSR\cite{msr_srivastava18_sltu} & 3 & - & 150 & Human & M & \cmark & \xmark & \xmark \\
%
OpenSLR \cite{openslr-kjartansson-etal-sltu2018} & 6 & - & 1247 & Human & M & \cmark & \xmark & \xmark \\
\begin{tabular}[c]{@{}l@{}}MSD \cite{cms-he-etal-2020-open}\end{tabular} & 6 & - & 35 & Human & M & \cmark & \xmark & \xmark \\
MUCS \cite{mucs-diwan2021multilingual} & 3 & - & 350 & Human & M & \cmark & \xmark & \xmark \\
IndicSUPERB \cite{kathbath-10.1609/aaai.v37i11.26521}& 12 & - & 1684 & Human & M & \cmark & \xmark & \xmark \\
Shrutilipi \cite{shrutilipi-10096933}& 12 & - & 6457 & Newsonair & FA & \cmark & \xmark & \xmark \\
Graamvaani \cite{gramvaani-DBLP:conf/interspeech/BhanushaliBGGKK22}& 1 & - & 108 & Human & M & \xmark & \xmark & \xmark \\
IIIS-Mile \cite{mile_1}& 2 & - & 500 & Human & M & \cmark & \xmark & \xmark \\
Vāksañcayah \cite{vakcan-adiga-etal-2021-automatic}& 1 & - & 78 & Human & M & \xmark & \xmark & \xmark \\
\begin{tabular}[c]{@{}l@{}}IIIT-H ISD \cite{iiit-isd-Prahallad2012TheII}\end{tabular} & 7 & - & 11 & Human & M & \cmark & \xmark & \xmark \\
\begin{tabular}[c]{@{}l@{}}MSR - IITB\cite{iitbmsc-marathidata}\end{tabular} & 1 & - & 109 & Human & M & \xmark & \xmark & \xmark \\
NPTEL \cite{nptel-Bhogale2023VistaarDB} & 8 & - & 6400 & YouTube & FA & \cmark & \xmark & \xmark \\
IndicTTS \cite{indic-tts-2016ResourcesFI}& 13 & - & 225 & Human & M & \cmark & \xmark & \xmark \\
Svarah \cite{svarah-javed23_interspeech}& 1 & 37 & 10 & Human & M & \xmark & \cmark & \xmark \\
SPRING-INX \cite{springx}& 10 & - & 3302 & Human & M & \cmark & \xmark & \xmark \\
SPIRE-SIES \cite{spiresies-singh2023spiresies}& 1 & 13 & 23 & Human & PL & \xmark & \cmark & \xmark \\
Lahaja \cite{lahaja}& 1 & 83 & 12.5 & Human & M & \xmark & \cmark & \xmark \\
\midrule
\textbf{Nirantar} & 22 & 208 & 3250 & Human & M & \cmark & \cmark & \cmark \\
\bottomrule
\end{tabular}
\label{tab:dataset_stats}

\end{table}
\endgroup

\subsection{Definitions}
\textbf{Data Batch ($B$):} 
A data batch, represented as an ordered tuple \( B = (l, d) \), is the outcome of a single data collection activity for a domain \( d \) of language \( l \), where \( l \in \mathcal{L} \) and \( d \in \mathcal{D} \). In ASR, each data batch comprises of $(x,y)$ pairs, where $x$ denotes the raw speech signal and $y$ represents the corresponding transcript.


\noindent\textbf{Episode ($E$):} An episode consists of one or more data batches (\( B \)) collected in parallel and is defined as a set of data batches:  $\label{eq:1} E = \{(l, d) \mid l \in \mathcal{L}, d \in \mathcal{D} \}$

\noindent\textbf{Timeline ($T$):} A timeline $T$ is defined as an ordered sequence of episodes $T = \langle E_0, .., E_t,.. , E_\tau \rangle$ where each $E_t$ represents an episode at time step $t$, and $\tau$ denotes the total no. of episodes.


\noindent\textbf{Model ($m$)}: A model $m$ is a learnt mapping $y = m(x)$ by training on a collection of data batches.

\noindent\textbf{Continual Learning Method ($c$):} Given a timeline $T$, and a base model $m_0$, the continual learning method $c(\cdot)$ produces a model $m_\tau$ iteratively: $m_{t} = c(E_t, m_{t-1}),  \quad  1 \leq t \leq \tau $

\subsection{Continual Learning Scenarios}
\label{subsub-timeline}

\noindent\textbf{Language Incremental Learning (LIL)}: In the LIL scenario, each episode introduces a new language. Specifically, at time step $t$, episode $E_t$ consists of all data batches associated with language $L_t$, i.e.,  $ E_t = \{ (L_t,d) \mid d \in \mathcal{D} \} \quad \forall\ t \in \tau, L_t \in \mathcal{L}$


\noindent\textbf{Domain Incremental Learning (DIL)}: In this scenario, all languages ($\mathcal{L}$) are introduced in base episode $E_0 = \{(l,d) | \cup l = \mathcal{L}\}$. In subsequent episodes $E_{t}$ where $1 \leq t \leq \tau$, only new domains are added, while the set of languages remains unchanged.

\noindent\textbf{Language-Incremental Domain-Incremental Learning (LIDIL)}: In this scenario, both new languages and new districts are introduced over time ($E_0$ to $E_{\tau}$). Episodes are formed by arbitrary collections of batches, and any sequence of these episodes forms a timeline.

\subsection{Dataset Description}

\begin{figure}[!t]
    \centering
    \includegraphics[width=\linewidth]{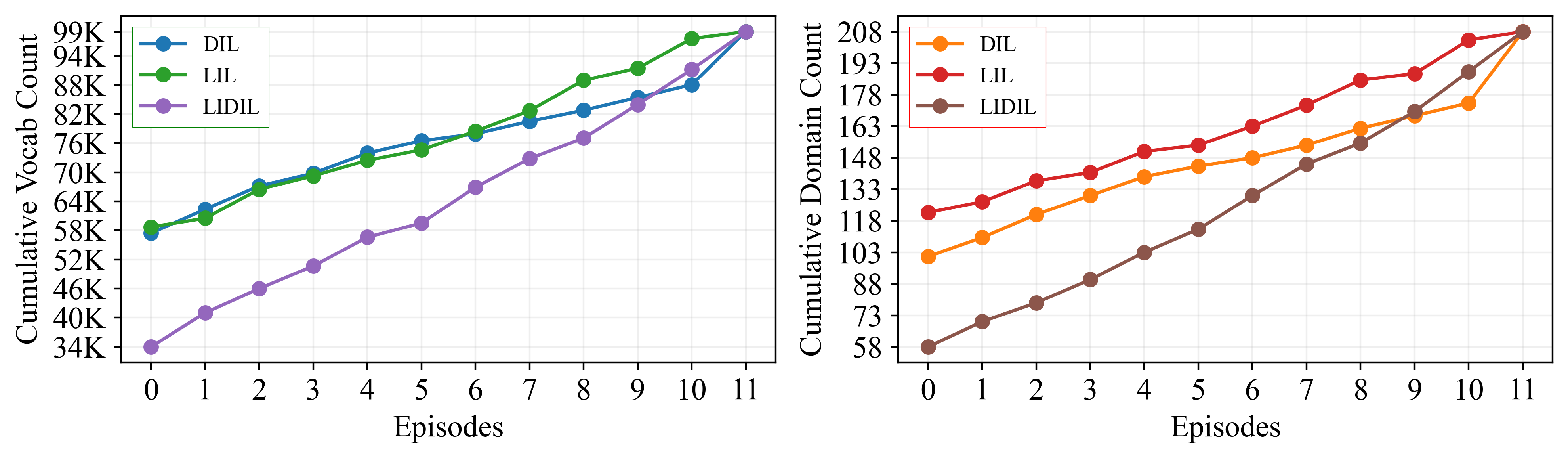}
    \caption{Evolution of vocabulary and domains across episodes}
    \label{fig:cumulative-evolution}
\end{figure}

Expanding on the IndicVoices\cite{indicvoices} effort, we introduce \dataset{}, designed for training and evaluating ASR systems in a continual learning (CL) setting. In addition to the initial 1530 hours from IndicVoices, we collect an additional 1720 hours using the same procedure covering a total of 22 languages and 208 districts. The data includes read, extempore, and conversational speech from diverse speakers, ensuring fair representation across age groups, genders, educational backgrounds, locations, and occupations. Data collection occurred in phases, with each phase covering one or more languages from different districts. Local coordinators mobilized 100-150 participants per district, obtaining consent and compensating them for their time. Participants engaged in three tasks: answering tailored questions on multiple domains and topics of interest, simulating voice assistant interactions, and engaging in two-party telephony conversations. Data was transcribed by an in-house team following a rigorous quality control process. Each district’s data forms a batch, and multiple batches aggregate into episodes, introducing variations in accents, vocabulary, and conversational topics. \dataset{} thus leverages the natural influx of audio data in batches and splices the audio speech data across multiple timelines, one each for LIL, DIL, LIDIL. The creation of the timelines is highlighted in Section \ref{subsec:timelines}. Table \ref{tab:dataset-stats} presents the statistics of data across languages. Figure \ref{fig:cumulative-evolution} shows the cumulative evolution of vocabulary and domains in Nirantar. For creating the test data, we sample a maximum of 15 minutes from each of the domains resulting in a total of 50 hours across languages. Since the test data contains samples from every district, we can evaluate the forward and backward transfer of CL approaches.  

\subsection{Continual Learning Playground}
\label{subsec:timelines}
Nirantar comprises of three distinct timelines corresponding to LIL, DIL and LIDIL scenarios respectively (see Table \ref{tab:playground_stats}). 
Next, we present the process of creation of the timelines.

\noindent\textbf{Base episode ($E_0$)}: 
In a practical scenario, the base model ($m_0$) will be trained after a seed amount of data is collected. We consider a good starting point for the base episode ($E_0$) when data batches are collected for half of the languages and half of the domains in each language. 
Specifically, for LIDIL, we select the 11 languages with the most hours in Table \ref{tab:dataset-stats}, and sample half their domains to create $E_0$. For LIL, we start with the same set of 11 languages having all domains of the respective languages. For DIL, we start with all 22 languages, and randomly sample half of the number of domains in each language.

\noindent\textbf{Incremental episodes ($E_{\tau \geq 1}$)}: We create timelines with $\tau=11$.  In LIL, each episode adds \textit{all} data batches of a single language, with languages introduced in random order.
For DIL and LIDIL, data batches are randomly assigned to episodes, ensuring uniform distribution of data batches while still maintaining non-uniformity in training hours across episodes (see Table \ref{tab:playground_stats}). In DIL, all languages appear in Episode 0, whereas in LIDIL, only half do, enabling incremental addition of both languages and domains in subsequent episodes. 


\noindent Given this playground, our goal is to identify an optimal continual learning approach $c^*$ for a given timeline $T$ and base model $m_0$. Formally $c^* = \min_{c\in \mathcal{C}} V(c \mid T, m)$, where $V$ is a metric that evaluates the approach, and $\mathcal{C}$ is a set of CL approaches.
 

\begingroup
\setlength{\tabcolsep}{3.5pt} 
\begin{table}[t]
    \scriptsize
    \centering
    \caption{Number of hours (\#H), speakers (\#Sp), and domains (\#D) in \dataset{}, along with the ISO codes for the languages.}
    \begin{tabular}{ll|rrr|@{}|@{}|ll|rrr}
    \toprule
    \textbf{Language} & \textbf{iso} & \textbf{\#H} & \textbf{\#Sp} & \textbf{\#D} & \textbf{Language} & \textbf{iso} & \textbf{\#H} & \textbf{\#Sp} & \textbf{\#D} \\
    \midrule
   \textbf{Assamese} & as & 241 & 985 & 14 & \textbf{Manipuri} & mni & 42 & 166 & 3 \\
   \textbf{Bengali} & bn & 209 & 733 & 11 & \textbf{Marathi} & mr & 118 & 447 & 10 \\
   \textbf{Bodo} & brx & 291 & 1061 & 4 & \textbf{Nepali} & ne & 252 & 780 & 4 \\
   \textbf{Dogri} & doi & 116 & 495 & 5 & \textbf{Odia} & or & 124 & 473 & 9 \\
   \textbf{Gujarati} & gu & 20 & 72 & 4 & \textbf{Punjabi} & pa & 124 & 344 & 6 \\
   \textbf{Hindi} & hi & 138 & 490 & 12 & \textbf{Sanskrit} & sa & 70 & 222 & 17 \\
   \textbf{Kannada} & kn & 96 & 530 & 13 & \textbf{Santali} & sat & 164 & 433 & 8 \\
   \textbf{Konkani} & kok & 103 & 245 & 4 & \textbf{Sindhi} & sd & 27 & 240 & 4 \\
   \textbf{Kashmiri} & ks & 106 & 515 & 10 & \textbf{Tamil} & ta & 238 & 1242 & 19 \\
   \textbf{Maithili} & mai & 248 & 726 & 9 & \textbf{Telugu} & te & 221 & 767 & 28 \\
   \textbf{Malayalam} & ml & 170 & 504 & 10 & \textbf{Urdu} & ur & 124 & 564 & 10 \\
   \bottomrule
   \end{tabular}
    \label{tab:dataset-stats}
\end{table}
\endgroup

\begingroup
\setlength{\tabcolsep}{1.26pt} 
\begin{table}[!t]
\tiny
\centering
\caption{Statistics showing district counts per language in LIL, DIL and LIDIL scenarios. Each cell carries the number of districts added for a given language (row) at episode (Ep).}
\begin{tabular}{l|@{}|cccccccccccc|@{}|cccccccccccc|@{}|cccccccccccc}
 \toprule
  & \multicolumn{12}{c|@{}|}{\textbf{LIL}} & \multicolumn{12}{c|@{}|}{\textbf{DIL}} & \multicolumn{12}{c}{\textbf{LIDIL}} \\
 \cmidrule{2-37}
 \textbf{Ep} & {\textbf{0}} & {\textbf{1}} & {\textbf{2}} & {\textbf{3}} & {\textbf{4}} & {\textbf{5}} & {\textbf{6}} & {\textbf{7}} & {\textbf{8}} & {\textbf{9}} & {\textbf{10}} & {\textbf{11}} & {\textbf{0}} & {\textbf{1}} & {\textbf{2}} & {\textbf{3}} & {\textbf{4}} & {\textbf{5}} & {\textbf{6}} & {\textbf{7}} & {\textbf{8}} & {\textbf{9}} & {\textbf{10}} & {\textbf{11}} & {\textbf{0}} & {\textbf{1}} & {\textbf{2}} & {\textbf{3}} & {\textbf{4}} & {\textbf{5}} & {\textbf{6}} & {\textbf{7}} & {\textbf{8}} & {\textbf{9}} & {\textbf{10}} & {\textbf{11}} \\
 \midrule
\textbf{as} & \cellcolor[HTML]{A9CD99}14 & - & - & - & - & - & - & - & - & - & - & - & \cellcolor[HTML]{F2BF89}7 & - & {\cellcolor[HTML]{FCE5CD}1} & {\cellcolor[HTML]{FCE5CD}1} & {\cellcolor[HTML]{FCE5CD}1} & - & {\cellcolor[HTML]{FCE5CD}1} & - & {\cellcolor[HTML]{FCE5CD}1} & - & - & \cellcolor[HTML]{FBDFC2}2 & \cellcolor[HTML]{8CB8DF}7 & - & - & {\cellcolor[HTML]{CFE2F3}1} & - & - & \cellcolor[HTML]{C4DBF0}2 & - & - & {\cellcolor[HTML]{CFE2F3}1} & \cellcolor[HTML]{B9D4ED}3 & - \\
\textbf{bn} & \cellcolor[HTML]{B6D5A9}11 & - & - & - & - & - & - & - & - & - & - & - & \cellcolor[HTML]{F6CCA0}5 & - & - & - & - & - & - & {\cellcolor[HTML]{FCE5CD}1} & - & {\cellcolor[HTML]{FCE5CD}1} & - & \cellcolor[HTML]{F7D2AB}4 & \cellcolor[HTML]{A3C6E6}5 & - & - & - & {\cellcolor[HTML]{CFE2F3}1} & {\cellcolor[HTML]{CFE2F3}1} & \cellcolor[HTML]{CFE2F3}1 & \cellcolor[HTML]{C4DBF0}2 & - & - & - & \cellcolor[HTML]{CFE2F3}1 \\
\textbf{brx} & \cellcolor[HTML]{D5E8CE}4 & - & - & - & - & - & - & - & - & - & - & - & \cellcolor[HTML]{FBDFC2}2 & - & - & - & - & {\cellcolor[HTML]{FCE5CD}1} & - & - & - & {\cellcolor[HTML]{FCE5CD}1} & - & - & \cellcolor[HTML]{C4DBF0}2 & - & - & - & - & - & \cellcolor[HTML]{CFE2F3}1 & - & - & - & - & \cellcolor[HTML]{CFE2F3}1 \\
\textbf{doi} & - & {\cellcolor[HTML]{D1E5C9}5} & - & - & - & - & - & - & - & - & - & - & \cellcolor[HTML]{FBDFC2}2 & {\cellcolor[HTML]{FCE5CD}1} & - & {\cellcolor[HTML]{FCE5CD}1} & - & - & - & {\cellcolor[HTML]{FCE5CD}1} & - & - & - & - & - & - & {\cellcolor[HTML]{C4DBF0}2} & - & - & - & - & \cellcolor[HTML]{C4DBF0}2 & - & - & - & \cellcolor[HTML]{CFE2F3}1 \\
\textbf{gu} & - & - & - & {\cellcolor[HTML]{D5E8CE}4} & - & - & - & - & - & - & - & - & \cellcolor[HTML]{FBDFC2}2 & - & - & - & {\cellcolor[HTML]{FCE5CD}1} & {\cellcolor[HTML]{FCE5CD}1} & - & - & - & - & - & - & - & - & - & - & - & - & \cellcolor[HTML]{CFE2F3}1 & \cellcolor[HTML]{CFE2F3}1 & - & {\cellcolor[HTML]{CFE2F3}1} & - & \cellcolor[HTML]{CFE2F3}1 \\
\textbf{hi} & \cellcolor[HTML]{B2D3A4}12 & - & - & - & - & - & - & - & - & - & - & - & \cellcolor[HTML]{F4C594}6 & - & - & {\cellcolor[HTML]{FCE5CD}1} & {\cellcolor[HTML]{FCE5CD}1} & - & - & - & {\cellcolor[HTML]{FBDFC2}2} & - & - & \cellcolor[HTML]{FBDFC2}2 & \cellcolor[HTML]{97BFE2}6 & {\cellcolor[HTML]{CFE2F3}1} & - & - & - & {\cellcolor[HTML]{CFE2F3}1} & \cellcolor[HTML]{CFE2F3}1 & \cellcolor[HTML]{CFE2F3}1 & - & - & \cellcolor[HTML]{CFE2F3}1 & \cellcolor[HTML]{CFE2F3}1 \\
\textbf{kn} & - & - & - & - & - & - & - & - & {\cellcolor[HTML]{ADD09F}13} & - & - & - & \cellcolor[HTML]{F4C594}6 & {\cellcolor[HTML]{F9D9B7}3} & - & - & - & {\cellcolor[HTML]{FCE5CD}1} & - & - & {\cellcolor[HTML]{FCE5CD}1} & - & {\cellcolor[HTML]{FCE5CD}1} & \cellcolor[HTML]{FCE5CD}1 & - & - & {\cellcolor[HTML]{C4DBF0}2} & {\cellcolor[HTML]{CFE2F3}1} & {\cellcolor[HTML]{C4DBF0}2} & {\cellcolor[HTML]{CFE2F3}1} & \cellcolor[HTML]{CFE2F3}1 & - & {\cellcolor[HTML]{C4DBF0}2} & {\cellcolor[HTML]{CFE2F3}1} & \cellcolor[HTML]{C4DBF0}2 & \cellcolor[HTML]{CFE2F3}1 \\
\textbf{kok} & - & - & - & - & - & {\cellcolor[HTML]{D5E8CE}4} & - & - & - & - & - & - & \cellcolor[HTML]{FBDFC2}2 & - & - & - & - & - & - & {\cellcolor[HTML]{FCE5CD}1} & - & - & {\cellcolor[HTML]{FCE5CD}1} & - & - & - & - & - & - & - & - & - & - & {\cellcolor[HTML]{CFE2F3}1} & \cellcolor[HTML]{C4DBF0}2 & \cellcolor[HTML]{CFE2F3}1 \\
\textbf{ks} & - & - & {\cellcolor[HTML]{BAD8AF}10} & - & - & - & - & - & - & - & - & - & \cellcolor[HTML]{F6CCA0}5 & {\cellcolor[HTML]{FCE5CD}1} & {\cellcolor[HTML]{FCE5CD}1} & - & - & - & {\cellcolor[HTML]{FCE5CD}1} & {\cellcolor[HTML]{FCE5CD}1} & - & {\cellcolor[HTML]{FCE5CD}1} & - & - & - & {\cellcolor[HTML]{B9D4ED}3} & {\cellcolor[HTML]{CFE2F3}1} & - & {\cellcolor[HTML]{CFE2F3}1} & - & - & \cellcolor[HTML]{CFE2F3}1 & - & - & \cellcolor[HTML]{CFE2F3}1 & \cellcolor[HTML]{B9D4ED}3 \\
\textbf{mai} & \cellcolor[HTML]{BFDBB4}9 & - & - & - & - & - & - & - & - & - & - & - & \cellcolor[HTML]{F7D2AB}4 & - & {\cellcolor[HTML]{FCE5CD}1} & - & {\cellcolor[HTML]{FCE5CD}1} & - & - & - & - & - & - & \cellcolor[HTML]{F9D9B7}3 & \cellcolor[HTML]{AECDE9}4 & - & {\cellcolor[HTML]{CFE2F3}1} & - & {\cellcolor[HTML]{C4DBF0}2} & {\cellcolor[HTML]{CFE2F3}1} & - & - & {\cellcolor[HTML]{CFE2F3}1} & - & - & - \\
\textbf{ml} & \cellcolor[HTML]{BAD8AF}10 & - & - & - & - & - & - & - & - & - & - & - & \cellcolor[HTML]{F6CCA0}5 & {\cellcolor[HTML]{FCE5CD}1} & {\cellcolor[HTML]{FCE5CD}1} & {\cellcolor[HTML]{FCE5CD}1} & - & - & - & - & - & {\cellcolor[HTML]{FCE5CD}1} & {\cellcolor[HTML]{FCE5CD}1} & - & \cellcolor[HTML]{A3C6E6}5 & {\cellcolor[HTML]{CFE2F3}1} & - & - & - & - & \cellcolor[HTML]{C4DBF0}2 & - & - & - & \cellcolor[HTML]{C4DBF0}2 & - \\
\textbf{mni} & - & - & - & - & - & - & - & - & - & {\cellcolor[HTML]{D9EAD3}3} & - & - & \cellcolor[HTML]{FCE5CD}1 & - & - & - & {\cellcolor[HTML]{FCE5CD}1} & - & - & - & - & - & - & \cellcolor[HTML]{FCE5CD}1 & - & - & - & - & {\cellcolor[HTML]{CFE2F3}1} & - & - & \cellcolor[HTML]{CFE2F3}1 & - & - & \cellcolor[HTML]{CFE2F3}1 & - \\
\textbf{mr} & - & - & - & - & - & - & - & {\cellcolor[HTML]{BAD8AF}10} & - & - & - & - & \cellcolor[HTML]{F6CCA0}5 & {\cellcolor[HTML]{FCE5CD}1} & {\cellcolor[HTML]{FBDFC2}2} & - & - & - & - & - & - & - & {\cellcolor[HTML]{FCE5CD}1} & \cellcolor[HTML]{FCE5CD}1 & - & {\cellcolor[HTML]{CFE2F3}1} & - & {\cellcolor[HTML]{CFE2F3}1} & {\cellcolor[HTML]{CFE2F3}1} & {\cellcolor[HTML]{CFE2F3}1} & - & - & - & {\cellcolor[HTML]{AECDE9}4} & \cellcolor[HTML]{CFE2F3}1 & \cellcolor[HTML]{CFE2F3}1 \\
\textbf{ne} & \cellcolor[HTML]{D5E8CE}4 & - & - & - & - & - & - & - & - & - & - & - & \cellcolor[HTML]{FBDFC2}2 & {\cellcolor[HTML]{FCE5CD}1} & {\cellcolor[HTML]{FCE5CD}1} & - & - & - & - & - & - & - & - & - & \cellcolor[HTML]{C4DBF0}2 & - & - & {\cellcolor[HTML]{CFE2F3}1} & - & - & - & - & - & {\cellcolor[HTML]{CFE2F3}1} & - & - \\
\textbf{or} & - & - & - & - & - & - & {\cellcolor[HTML]{BFDBB4}9} & - & - & - & - & - & \cellcolor[HTML]{F7D2AB}4 & - & {\cellcolor[HTML]{FCE5CD}1} & {\cellcolor[HTML]{FCE5CD}1} & - & - & - & - & - & - & - & \cellcolor[HTML]{F9D9B7}3 & - & {\cellcolor[HTML]{C4DBF0}2} & - & - & - & - & - & - & {\cellcolor[HTML]{B9D4ED}3} & {\cellcolor[HTML]{C4DBF0}2} & \cellcolor[HTML]{CFE2F3}1 & \cellcolor[HTML]{CFE2F3}1 \\
\textbf{pa} & \cellcolor[HTML]{CCE3C4}6 & - & - & - & - & - & - & - & - & - & - & - & \cellcolor[HTML]{F9D9B7}3 & - & - & - & - & - & - & {\cellcolor[HTML]{FBDFC2}2} & {\cellcolor[HTML]{FCE5CD}1} & - & - & - & \cellcolor[HTML]{B9D4ED}3 & - & - & - & - & - & \cellcolor[HTML]{CFE2F3}1 & - & - & {\cellcolor[HTML]{CFE2F3}1} & - & \cellcolor[HTML]{CFE2F3}1 \\
\textbf{sa} & - & - & - & - & - & - & - & - & - & - & {\cellcolor[HTML]{9BC68A}17} & - & \cellcolor[HTML]{F1B87D}8 & - & - & - & - & - & - & - & - & {\cellcolor[HTML]{FCE5CD}1} & {\cellcolor[HTML]{FCE5CD}1} & \cellcolor[HTML]{F2BF89}7 & - & {\cellcolor[HTML]{C4DBF0}2} & {\cellcolor[HTML]{C4DBF0}2} & {\cellcolor[HTML]{B9D4ED}3} & {\cellcolor[HTML]{CFE2F3}1} & {\cellcolor[HTML]{B9D4ED}3} & \cellcolor[HTML]{CFE2F3}1 & - & {\cellcolor[HTML]{CFE2F3}1} & {\cellcolor[HTML]{C4DBF0}2} & \cellcolor[HTML]{CFE2F3}1 & \cellcolor[HTML]{CFE2F3}1 \\
\textbf{sat} & \cellcolor[HTML]{C3DDB9}8 & - & - & - & - & - & - & - & - & - & - & - & \cellcolor[HTML]{F7D2AB}4 & - & - & - & - & - & - & - & - & - & - & \cellcolor[HTML]{F7D2AB}4 & \cellcolor[HTML]{AECDE9}4 & {\cellcolor[HTML]{CFE2F3}1} & - & - & - & - & \cellcolor[HTML]{CFE2F3}1 & \cellcolor[HTML]{CFE2F3}1 & - & - & \cellcolor[HTML]{CFE2F3}1 & - \\
\textbf{sd} & - & - & - & - & - & - & - & - & - & - & - & {\cellcolor[HTML]{D5E8CE}4} & \cellcolor[HTML]{FBDFC2}2 & - & {\cellcolor[HTML]{FBDFC2}2} & - & - & - & - & - & - & - & - & - & - & {\cellcolor[HTML]{CFE2F3}1} & - & {\cellcolor[HTML]{CFE2F3}1} & - & {\cellcolor[HTML]{CFE2F3}1} & - & \cellcolor[HTML]{CFE2F3}1 & - & - & - & - \\
\textbf{ta} & \cellcolor[HTML]{92C07F}19 & - & - & - & - & - & - & - & - & - & - & - & \cellcolor[HTML]{EFB272}9 & {\cellcolor[HTML]{FCE5CD}1} & - & {\cellcolor[HTML]{FCE5CD}1} & {\cellcolor[HTML]{FCE5CD}1} & {\cellcolor[HTML]{FCE5CD}1} & - & - & {\cellcolor[HTML]{FCE5CD}1} & - & {\cellcolor[HTML]{FBDFC2}2} & \cellcolor[HTML]{F9D9B7}3 & \cellcolor[HTML]{76A9D8}9 & - & {\cellcolor[HTML]{CFE2F3}1} & - & {\cellcolor[HTML]{CFE2F3}1} & - & \cellcolor[HTML]{C4DBF0}2 & \cellcolor[HTML]{CFE2F3}1 & {\cellcolor[HTML]{CFE2F3}1} & - & - & \cellcolor[HTML]{AECDE9}4 \\
\textbf{te} & \cellcolor[HTML]{6AA84F}28 & - & - & - & - & - & - & - & - & - & - & - & \cellcolor[HTML]{E69138}14 & {\cellcolor[HTML]{FCE5CD}1} & {\cellcolor[HTML]{FCE5CD}1} & {\cellcolor[HTML]{F9D9B7}3} & {\cellcolor[HTML]{F9D9B7}3} & - & {\cellcolor[HTML]{FCE5CD}1} & - & {\cellcolor[HTML]{FCE5CD}1} & {\cellcolor[HTML]{FCE5CD}1} & - & \cellcolor[HTML]{F9D9B7}3 & \cellcolor[HTML]{3D85C6}14 & - & - & {\cellcolor[HTML]{C4DBF0}2} & {\cellcolor[HTML]{B9D4ED}3} & {\cellcolor[HTML]{C4DBF0}2} & \cellcolor[HTML]{CFE2F3}1 & \cellcolor[HTML]{C4DBF0}2 & {\cellcolor[HTML]{CFE2F3}1} & {\cellcolor[HTML]{CFE2F3}1} & \cellcolor[HTML]{C4DBF0}2 & - \\
\textbf{ur} & - & - & - & - & {\cellcolor[HTML]{BAD8AF}10} & - & - & - & - & - & - & - & \cellcolor[HTML]{F6CCA0}5 & - & - & - & - & {\cellcolor[HTML]{FCE5CD}1} & {\cellcolor[HTML]{FCE5CD}1} & - & {\cellcolor[HTML]{FCE5CD}1} & - & - & \cellcolor[HTML]{FBDFC2}2 & - & - & - & {\cellcolor[HTML]{CFE2F3}1} & {\cellcolor[HTML]{CFE2F3}1} & - & \cellcolor[HTML]{CFE2F3}1 & \cellcolor[HTML]{C4DBF0}2 & {\cellcolor[HTML]{CFE2F3}1} & - & \cellcolor[HTML]{B9D4ED}3 & \cellcolor[HTML]{CFE2F3}1 \\
\bottomrule
\end{tabular}

\label{tab:playground_stats}

\end{table}
\endgroup
\section{Experimental setup}

\subsection{Continual Learning Methods}
Below, we list down all the approaches considered in this work.


\noindent\textbf{Incremental Finetuning (Inc. FT):} Given a base model $m_0$, we sequentially finetune models $m_{1 \leq t \leq \tau}$ using the data batches in $E_t$, and initializing the weights of $m_t$ using the trained $m_{t-1}$.

\noindent\textbf{Joint Finetuning (Joint FT):} Similar to Inc. FT, we sequentially finetune $m_{1 \leq t \leq \tau}$ by initializing $m_t$'s weights using the trained model $m_{t-1}$, but take the data batches from $\bigcup_{i=0}^{t}\{E_i\}$.

\noindent\textbf{Elastic Weight Consolidation (EWC) \cite{DBLP:conf/aaai/LiuYW21}:} EWC preserves important parameters from previous episodes by estimating parameter importance using the Fisher information matrix (F) and adds a penalty term to the loss function during training on the current task. This penalty term, controlled by hyperparameters $\lambda$ and $\alpha$, balances between adapting to new tasks and retaining old knowledge. Following \cite{DBLP:journals/corr/abs-2310-16931} we set $\lambda$ to 5 and $\alpha$ to 0.5.

\noindent\textbf{Experience Replay (ER) \cite{experience-replay}:} Experience replay stores data from previous episodes in a memory buffer and replays them during the training of models on current episodes. Following \cite{DBLP:journals/corr/abs-2310-16931}, we sample 3\% of data across each episode.

\noindent\textbf{Memory-aware Synapse (MAS) \cite{memory-aware-synapse}:} Like EWC, this method confines large model updates to weights. However, unlike the Fisher information matrix, it assesses parameter importance using the average magnitude of gradients of the squared L2 norm of the learned function. Following \cite{DBLP:journals/corr/abs-2310-16931}, we set $\alpha$ and $\lambda$ to 1 and 0.5, respectively. These values determine the relative strength of the regularization term and the influence of previous tasks on updating parameter importance.

\noindent\textbf{Architecture based methods \cite{DBLP:conf/icassp/Eeckth23a}:} 
These methods add parameters to the backbone network, but are unsuitable for DIL and LIDIL because the complexity grows as the number of episodes increases, with new parameters required for each language (22) and domain (208). Hence, we only use adapters for the LIL setup where we added up to 11 adapters (one for each new language). Adapters with a bottleneck dimension of 64 were added to each Conformer block of the Conformer-L model, introducing an extra 1 million parameters per language. 

\subsection{Training}
\label{subsec:training}
We train Conformer-L \cite{conformer} models, consisting of 120M parameters, as the encoder, with a hybrid CTC-RNNT \cite{ctc-rnnt} decoder. The model has 17 conformer blocks with 512 as the model dimension. The output vocabulary is of size 256 per language, and is created by a Byte-Pair-Encoding (BPE) tokenizer. Each language consists of a separate decoder head. All our models are trained using the Nvidia's NeMo library. The base models $m_0$ and the Joint FT models were trained for 150,000 steps with a constant learning rate of 0.0001. Due to the skew in data distribution across languages in our joint multilingual setup, we follow existing works \cite{temp_sampling2} and use  temperature sampling for better convergence. We trained the incremental models for 30K steps with half the learning rate. We trained the models using Adam optimizer with effective batch size of 8 audios per GPU. 

\subsection{Metrics}
To evaluate different CL strategies, we use the following standard metrics commonly used in CL literature \cite{DBLP:journals/corr/abs-2310-16931}. However we use MER \cite{mer} instead of WER, as MER is bounded between 0 to 1 and thus ensures a more standardised evaluation.

\noindent \textbf{AMER}: Calculates the average Match Error Rate (MER) across all seen episodes.
$AMER_{t} = \frac{1}{t} \sum_{i=1}^{t} MER_{t,i} \quad; t \in [0,\tau]$

\noindent \textbf{Forward Transfer (FWT)}: Captures how well the model leverages past knowledge to improve performance on new episodes. 
$FWT_{t} = MER_t^{inc. ft} - MER_{t,t}$; where $MER_t^{inc. ft}$ refers to the MER obtained from the model trained on episode $E_t$.

\noindent \textbf{Backward Transfer (BWT)}: Measures the effect of learning new tasks over the prior ones: negative values signal forgetting, while positive values indicate knowledge reinforcement.\\
$BWT_{t} = \frac{1}{t-1} \sum_{i=1}^{t-1} MER_{i,i} - MER_{t,i} \quad; t \in [1,\tau]$

\noindent \textbf{Intransigence Measure (IM)}: Evaluates the model’s ability to learn new tasks effectively, reflecting its plasticity. $IM_{t} = MER_{t,t} - MER_{t}^{joint ft}$ where $MER_{t}^{joint ft}$ is the MER of the model trained jointly on episodes ${\{E_0,..,E_t}\}$.






\section{Results and Discussions}



\begin{figure}[t]
    \centering
    \includegraphics[width=\linewidth]{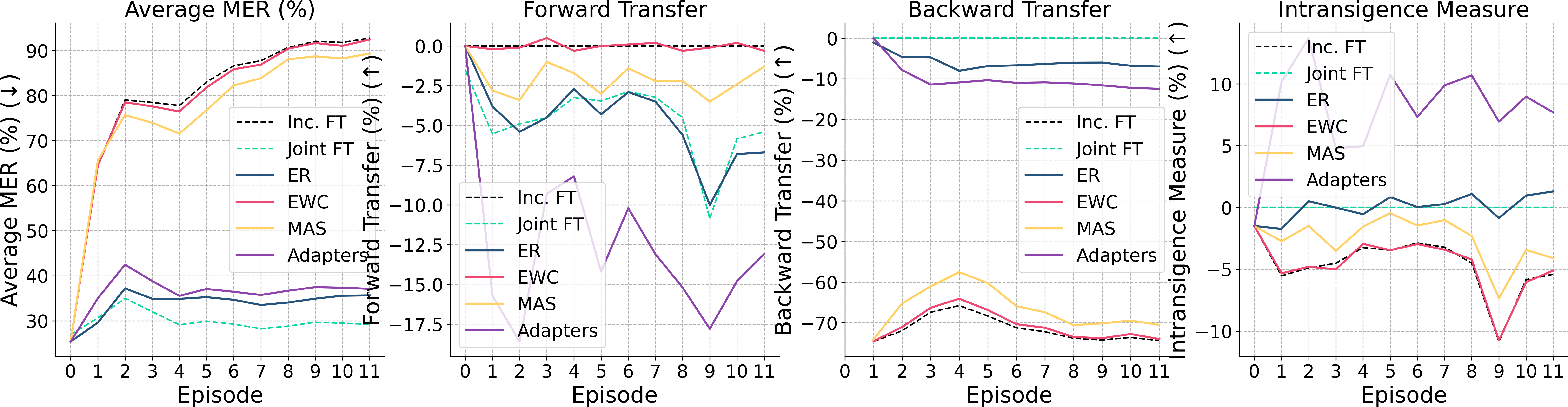}

    \centering
    \includegraphics[width=\columnwidth]{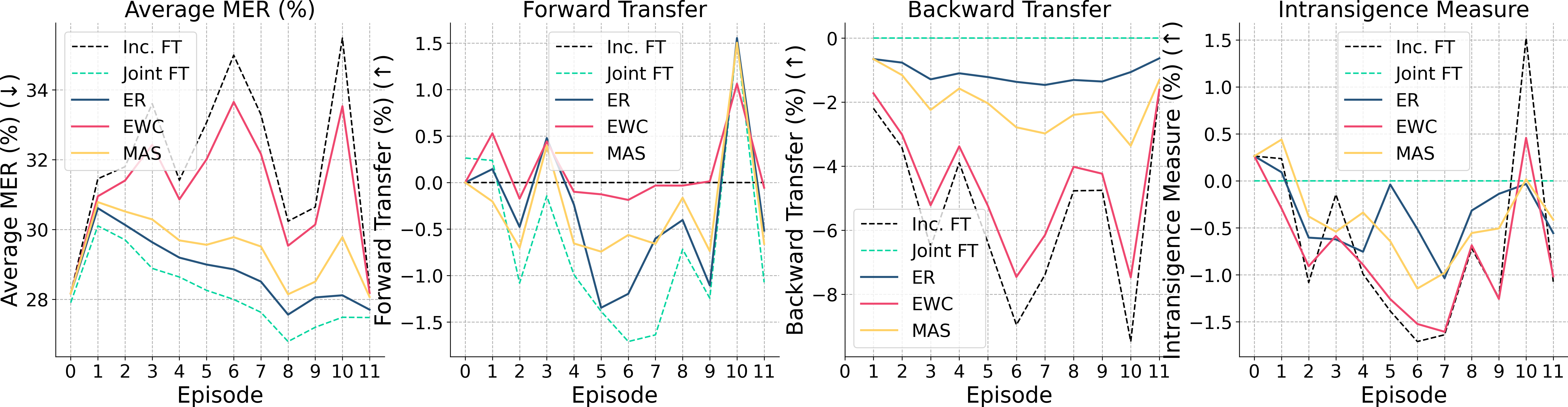}

    \centering
    \includegraphics[width=\linewidth]{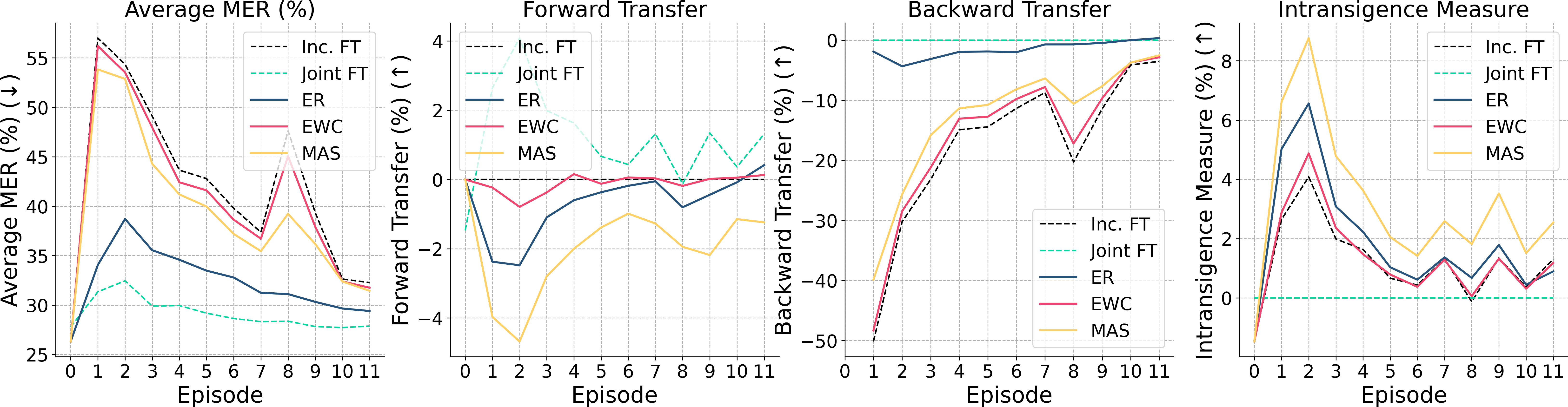}
    \caption{Comparison of various CL methods: \textbf{(top)} \textit{Language Incremental Learning (LIL)}, \textbf{(middle)} \textit{Domain Incremental Learning (LIL)} and \textbf{(bottom)} \textit{Language-Incremental Domain-Incremental Learning (LIDIL)}}
    \label{fig:lidil-dil-lil}
\end{figure}

\textbf{LIL:} Referring to Figure \ref{fig:lidil-dil-lil} (top), we observe a steady increase in AMER as new languages are introduced for Incremental FT, which is undesirable. 
Both regularization-based approaches, EWC and MAS, struggle to retain knowledge of previously learned languages, as shown by the trends in the Forward Transfer (FWT) across episodes. In contrast, ER significantly outperforms them, even with a buffer size of just 3\%, demonstrating the importance of replay in LIL. While ER demonstrates strong backward transfer (BWT) and positive intransigence, its poor forward transfer further emphasizes the need for CL approaches that better leverage knowledge from previous episodes. We also observe a sharp drop in the forward transfer and intransigence measures at episode 9. We hypothesize that this decline is due to the introduction of Manipuri, a Tibeto-Burman language with only 26 hours of data. The limited data and its notable differences from the Indo-Aryan and Dravidian language families observed in earlier episodes are likely factors contributing to this decline. Adapters outperform most CL approaches, except ER, in AMER and BWT by preventing forgetting with separate adapter layers per episode. However, their FWT is lower due to limited knowledge sharing, and their high Intransigence Measure and growing parameter count (11M by the final episode) make them impractical for large-scale incremental settings.


\textbf{DIL:} Referring to Figure \ref{fig:lidil-dil-lil} (middle), unlike LIL, we observe that AMER reduces over episodes for two methods, MAS and ER. The reduction of AMER over episodes could be attributed to (i) current CL methods being able to adapt better to new domains than to new languages, and (ii) the slightly favorable scenario in DIL, where the base model has already seen all the languages. 
All CL approaches demonstrate good forward transfer and intransigence measure in DIL.  The observed performance change of only 1.5\% is due to the randomness in the order of incoming data batches. This indicates that knowledge from previous domains is indeed helpful for new domains. 
Although MAS performs poorly in LIL, it shows good Forward and Backward Transfer in DIL, indicating that regularization-based methods are well-suited for domain-incremental learning.

\textbf{LIDIL:} In Figure \ref{fig:lidil-dil-lil} (bottom), we observe across all methods that the AMER first increases in the first 2 episodes similar to LIL, and then steadily decreases from episode 3 onwards, similar to DIL. This is due to the fact that many new languages are seen in the first 2 episodes, and the number of new languages gradually reduces after that. This demonstrates the unique hybrid nature of this newly introduced continual learning scenario that encompasses characteristics from both the aforementioned scenarios, \textit{viz.}, LIL and DIL. We also observe that backward transfer for EWC and MAS improves over time, unlike the other methods, indicating gradual adaptation to previous tasks as new languages and domains are added. All methods show a positive Intransigence Measure in LIDIL. Lastly, to verify impact of episode order, we tested three randomized sequences in the LIDIL scenario. Results showed consistent AMER and BWT scores, stable method rankings, and some variation in intransigence, suggesting certain episodic sequences are harder to train. Due to space constraints, these results are not included.



Our experiments thus demonstrate that no single method consistently excels across all three scenarios, underscoring the need for more robust CL approaches to  handle the real-world incremental learning challenges presented in Nirantar. 

\section{Conclusion}
We presented Nirantar, a novel data framework designed to facilitate training and evaluation of continual learning (CL) methods in multilingual and multidomain settings. This dataset contains 3250 hours of human-transcribed speech data, including 1720 hours released from this study, organized into 12 episodes featuring diverse language and domain combinations. Evaluations using established CL methods such as Elastic Weight Consolidation, Memory-aware Synapse, and Experience Replay highlight the utility of the dataset across Language-Incremental (LIL), Domain-Incremental (DIL), and Language-Incremental Domain-Incremental Learning (LIDIL) scenarios. 
All associated resources have been released\footref{fn:nirantar} under a CC-BY-4 license to support further research in this area.

\bibliographystyle{IEEEtran}
\bibliography{main}

\end{document}